\documentclass{article}

\usepackage{enumitem}
\usepackage{CJKutf8}
\usepackage{caption}
\usepackage{amsmath}
\usepackage{arxiv}

\usepackage[utf8]{inputenc} 
\usepackage[T1]{fontenc} 
\usepackage{hyperref} 
\usepackage{url}
\usepackage{tabularx} 
\usepackage{makecell} 
\usepackage{threeparttable} 
\usepackage{booktabs} 
\usepackage{amsfonts} 
\usepackage{nicefrac} 
\usepackage{microtype}
\usepackage{lipsum}
\usepackage{graphicx}
\graphicspath{ {./images/} }

\title{A Disease Labeler for Chinese Chest X-Ray Report Generation}

\author{
 Mengwei Wang\thanks{These authors contributed equally to this work.} \\
  School of Artificial Intelligence\\
  Beijing University of Posts and Telecommunications\\
  Beijing 100876, China \\
  \texttt{wangmengwei@bupt.edu.cn} \\
\And
 Ruixin Yan\footnotemark[1] \\
  Department of Radiology\\
  Peking University Third Hospital\\
  Beijing 100191, China \\
  \texttt{yanruixin1996@163.com} \\
  \And
 Zeyi Hou \\
  School of Artificial Intelligence\\
  Beijing University of Posts and Telecommunications\\
  Beijing 100876, China \\
  \texttt{houzeyi@bupt.edu.cn} \\
  \And
 Ning Lang \\
  Department of Radiology\\
  Peking University Third Hospital\\
  Beijing 100191, China \\
  \texttt{langning800129@126.com} \\
  \And
 Xiuzhuang Zhou \\
  School of Artificial Intelligence\\
  Beijing University of Posts and Telecommunications\\
  Beijing 100876, China \\
  \texttt{xiuzhuang.zhou@bupt.edu.cn} \\
}

\begin{document}
\begin{CJK}{UTF8}{gbsn}
\maketitle
\begin{abstract}
In the field of medical image analysis, the scarcity of Chinese chest X-ray report datasets has hindered the development of technology for generating Chinese chest X-ray reports. On one hand, the construction of a Chinese chest X-ray report dataset is limited by the time-consuming and costly process of accurate expert disease annotation. On the other hand, a single natural language generation metric is commonly used to evaluate the similarity between generated and ground-truth reports, while the clinical accuracy and effectiveness of the generated reports rely on an accurate disease labeler (classifier). To address the issues, this study proposes a disease labeler tailored for the generation of Chinese chest X-ray reports. This labeler leverages a dual BERT architecture to handle diagnostic reports and clinical information separately and constructs a hierarchical label learning algorithm based on the affiliation between diseases and body parts to enhance text classification performance. Utilizing this disease labeler, a Chinese chest X-ray report dataset comprising 51,262 report samples was established. Finally, experiments and analyses were conducted on a subset of expert-annotated Chinese chest X-ray reports, validating the effectiveness of the proposed disease labeler.
\end{abstract}


\section{Introduction}

In the modern healthcare system, efficient and precise medical imaging reports play a crucial role in patient diagnosis and treatment planning. With the rapid increase in medical imaging data, traditional manual interpretation methods, which are time-consuming and limited by the personal experience of physicians, struggle to meet the demands of the rapidly evolving medical field. Therefore, there is an urgent need to leverage deep learning technology to automate the generation of medical imaging reports\cite{allaouzi2018automatic}. The task of generating medical imaging reports driven by deep learning relies on large volumes of high-quality datasets. However, annotated medical data are extremely scarce, and for researchers working on automated report generation, these data are as precious as rare minerals. Most studies rely on publicly available datasets to develop and test models. Among various medical imaging modalities, chest X-ray are not only the most common but also have a vast amount of data; hence, existing public datasets primarily consist of chest X-ray reports\cite{demner2016preparing,johnson2019MIMIC}. Moreover, these datasets are mostly in English, and there is a significant lack of Chinese chest X-ray report datasets, which limits the research on Chinese chest X-ray report generation algorithms.     

The main challenge in constructing a Chinese chest X-ray report dataset is the accurate annotation of disease labels. Generally, a dataset comprises three parts: reports, images, and labels. Reports and images can be obtained through a hospital's Radiology Information System (RIS) and Picture Archiving and Communication System (PACS), but the process of annotating disease labels is both complex and time-consuming. Although crowdsourcing methods can be used for large-scale dataset annotation in other fields, in the medical domain, this approach is both costly and impractical due to the need for specialized knowledge. Researchers have found that medical images usually come with corresponding medical reports, which implicitly contain the disease diagnostic labels of the images. Therefore, it is possible to extract disease labels for images from medical diagnostic reports. To this end, numerous natural language processing systems have been developed to utilize medical domain knowledge and construct complex feature engineering to annotate reports, thereby assisting in the disease annotation work of medical images\cite{pons2016natural}. With the advancement of artificial intelligence technology, deep learning methods have also been applied to the disease annotation of chest X-ray datasets, showing significant effects\cite{Irvin2019chexpert,smit2020combining}. However, existing methods are primarily designed for English datasets, and there is a lack of disease labelers suitable for Chinese medical reports.    

On the other hand, when evaluating the generation model of chest X-ray reports, natural language generation (NLG) evaluation metrics are commonly used, which can only measure the similarity between the generated text and the reference text but cannot evaluate the accuracy of disease prediction in the generated report. To assess the disease prediction accuracy of automated chest X-ray report generation models, Liu et al.\cite{liu2019clinically} proposed the Clinical Efficacy (CE) evaluation metric, which evaluates the disease prediction effect by calculating the F1 score, precision, and recall rate of the diseases included in the generated report, offering greater clinical practical value. However, calculating the CE metric requires a disease labeler to annotate the disease labels in the generated report. While there are standard disease annotation tools in the English domain, the Chinese domain lacks corresponding tools, which hinders the evaluation of Chinese chest X-ray report generation models and, consequently, the research on Chinese chest X-ray report generation algorithms.   

To address the aforementioned issues, this paper proposes a disease labeler with a dual BERT architecture and hierarchical label learning, and constructs a Chinese chest X-ray report dataset based on this labeler. The main contributions of this paper can be summarized as follows:    

\begin{enumerate}[label=\arabic*),leftmargin=*]
\item A Chinese disease labeler combining a dual BERT architecture with hierarchical label learning algorithms has been designed and developed. The labeler encodes diagnostic reports and clinical information through the dual BERT architecture and leverages the hierarchical relationship between diseases and body parts to create a hierarchical label learning algorithm, enhancing the labeler's disease prediction accuracy.   

\item Based on this disease labeler, a method for constructing a Chinese chest X-ray report dataset has been proposed, and a Chinese chest X-ray report dataset (CCXRD) has been constructed, providing a standardized process for the construction of related datasets.  

\item Experiments conducted on a subset of Chinese data annotated by experts show that the disease labeler proposed in this study exhibits superior annotation performance compared to existing models. As a general-purpose tool, it can not only be used for the construction of Chinese datasets but also for evaluating the disease prediction accuracy of Chinese chest X-ray report generation models.   
\end{enumerate}

\section{Related Word}

\subsection{Chest X-Ray Report Disease Labeler}  

In radiological examinations, chest X-ray is a routine and widely used procedure. Disease labels are often obtained through rule-based methods applied to chest X-ray reports to construct some publicly available large-scale datasets\cite{johnson2019MIMIC}. These rule-based methods typically rely on features such as controlled lexicons and syntactic rules for engineering, in order to identify and categorize radiological findings. NegEx\cite{chapman2001a} is a widely used rule component that identifies negations in reports through simple regular expressions and is often applied in conjunction with ontologies such as the Unified Medical Language System (UMLS). NegBio\cite{peng2018negbio} is an extension of NegEx, which not only detects negations in chest X-ray reports but also identifies uncertainties in various pathological states by defining general dependency patterns and subgraph matching for graph search traversal. NegBio has been applied to generate labels for the ChestX-Ray14 dataset\cite{wang2017chest}. The CheXpert labeler\cite{Irvin2019chexpert} improves performance in disease annotation of chest X-ray reports through more refined mention extraction, an improved natural language processing (NLP) pipeline, and rules for uncertainty and negation extraction. It played a significant role in creating the CheXpert dataset\cite{Irvin2019chexpert} and the MIMIC-CXR dataset\cite{johnson2019MIMIC} labels, which are among the largest publicly available chest X-ray datasets. Despite the notable progress of these rule-based methods, they have not fully captured the complexity, ambiguity, and subtlety of natural language in radiology reports.   

With the continuous advancement of artificial intelligence technology, deep learning methods have also achieved significant results in annotating radiology report datasets\cite{xue2019fine}. For instance, Bustos et al.\cite{bustos2020padchest} trained an ensemble of attention-augmented recurrent neural networks (RNNs) and convolutional neural networks (CNNs) on 27,593 reports annotated by physicians and generated labels accordingly. Recently, Transformer-based models have also been applied to the annotation tasks of radiology reports. Drozdov et al.\cite{drozdov2020supervised} trained classifiers with BERT and XLNet on 3,856 radiologist-annotated reports to distinguish between normal and abnormal labels. McDermott et al.\cite{mcdermott2020chexpertpp} proposed CheXpert++, an improved version of CheXpert\cite{Irvin2019chexpert}, trained on a BERT model and fine-tuned on expert-annotated reports based on the output of the rule-based CheXpert labeler, demonstrating higher performance, speed, differentiability, and probabilistic output. Smit et al.\cite{smit2020combining} combined the strengths of existing radiology report labelers and expert annotations to create CheXbert, a highly accurate chest X-ray report labeler.      

However, the aforementioned research primarily targets disease labelers designed for English chest X-ray reports. This paper takes into account the characteristics of Chinese chest X-ray reports and designs a Chinese chest X-ray report disease labeler with a dual BERT architecture and hierarchical label learning algorithm.   

\subsection{Chest X-Ray Report Generation Datasets}

The performance of deep learning-based medical imaging report generation algorithms is greatly influenced by the quality of the training data. High-quality datasets are the cornerstone of training these algorithms, and the availability of public datasets has greatly facilitated progress in the field of medical imaging report generation. Currently, public datasets are mostly sourced from the United States and Europe, focusing on chest X-ray (CXR) images and their corresponding reports. Table \ref{tab:datasets} details these typical chest X-ray datasets, including Indiana University Chest X-ray (IU X-Ray)\cite{demner2016preparing}, ChestX-ray\cite{wang2017chest}, CX-CHR\cite{li2018hybrid}, CheXpert\cite{Irvin2019chexpert}, MIMIC-CXR\cite{johnson2019MIMIC}, PadChest\cite{bustos2020padchest}, and CC-CXRI\cite{wang2021a}, among others.  

\begin{table}[h]
\centering
\caption{Common Chest X-ray Image Report Datasets} 
\label{tab:datasets}
\begin{tabular}{cccc} 
\hline
\textbf{Dataset}  & \textbf{Year} & \textbf{Number of Images} & \textbf{Language} \\
\hline
IU X-Ray\cite{demner2016preparing}& 2015 & 7,470  & English \\
ChestX-ray\cite{wang2017chest} & 2017 & 112,120& English \\
CX-CHR\cite{li2018hybrid}  & 2018 & 45,598 & Chinese \\
CheXpert\cite{Irvin2019chexpert}& 2019 & 224,316& English \\
MIMIC-CXR\cite{johnson2019MIMIC}  & 2019 & 377,110& English \\ 
PadChest\cite{bustos2020padchest}& 2019 & 160,868& English \\ 
CC-CXRI\cite{wang2021a}& 2021 & 161,398& Chinese \\
\hline
\end{tabular}
\end{table}

Specifically, IU X-Ray\cite{demner2016preparing} is an open-access, large-scale CXR dataset maintained by the United States National Library of Medicine, containing 7,470 images of lateral and frontal views of CXRs along with 3,955 corresponding radiological reports. These images have a resolution of 512×512 pixels and are provided in PNG format. The ChestX-ray dataset\cite{wang2017chest}, maintained by the National Institutes of Health Clinical Center, is divided into two versions: ChestX-ray8 and ChestX-ray14. ChestX-ray8 includes 108,948 frontal view CXR images of 32,717 unique patients and annotates eight disease types from radiological reports. ChestX-ray14 adds six more categories of chest diseases and encompasses a total of 112,120 frontal view CXR images. These two versions of the dataset are commonly used for classification tasks of chest X-ray, as well as for pre-training visual models in chest X-ray report generation tasks. CX-CHR\cite{li2018hybrid} is a private internal dataset consisting of 45,598 chest X-ray images and corresponding Chinese reports from 35,609 patients, but the specifics of its dataset construction method and process are not elaborated in detail in the literature\cite{li2018hybrid}, and it is only used within that publication. CheXpert\cite{Irvin2019chexpert} is a public dataset collected by Stanford Hospital, covering 22,431 frontal and lateral view CXRs from 65,240 patients, annotated as positive, negative, or uncertain for 14 common chest diseases using the rule-based CheXpert labeler\cite{Irvin2019chexpert}. MIMIC-CXR\cite{johnson2019MIMIC} is the largest public radiology dataset to date, provided by the Beth Israel Deaconess Medical Center in Boston, comprising 377,110 CXR images and 227,835 corresponding radiological reports. PadChest\cite{bustos2020padchest} comes from the Hospital San Juan in Spain, a large public dataset containing over 160,000 CXR images from 67,000 patients. CC-CXRI\cite{wang2021a} was constructed by the Chinese Chest X-ray Image Research Consortium, a vast CXR dataset whose disease labels were extracted from related radiological reports using a rule-based Natural Language Processing (NLP) labeler.      

The aforementioned work primarily involves English chest X-ray report datasets and Chinese chest X-ray datasets that only include chest X-ray and disease labels. However, this study, based on the proposed disease labeler, designs a construction process for a Chinese chest X-ray report dataset and completes the construction of a dataset. It also lays the foundation for the construction of large-scale publicly available datasets.   

\section{Chinese Chest X-Ray Report Disease Labeler} 

\subsection{Dual BERT Network Architecture}  

The objective of chest X-ray report annotation tasks is to extract disease labels from the free text of chest X-ray reports, such as "肺纹理增多" or "肺间质性病变". Specifically, the labeler receives paragraphs from radiological reports as input and outputs the determination results for each predefined disease category, classifying them as either positive or negative. It is noteworthy that the category "未见明显异常" is mutually exclusive with other disease categories, meaning that if any disease is detected, "未见明显异常" is classified as negative, and vice versa if no disease is detected.     

\begin{figure}[htb]
\centering 
\includegraphics[width=0.8\textwidth]{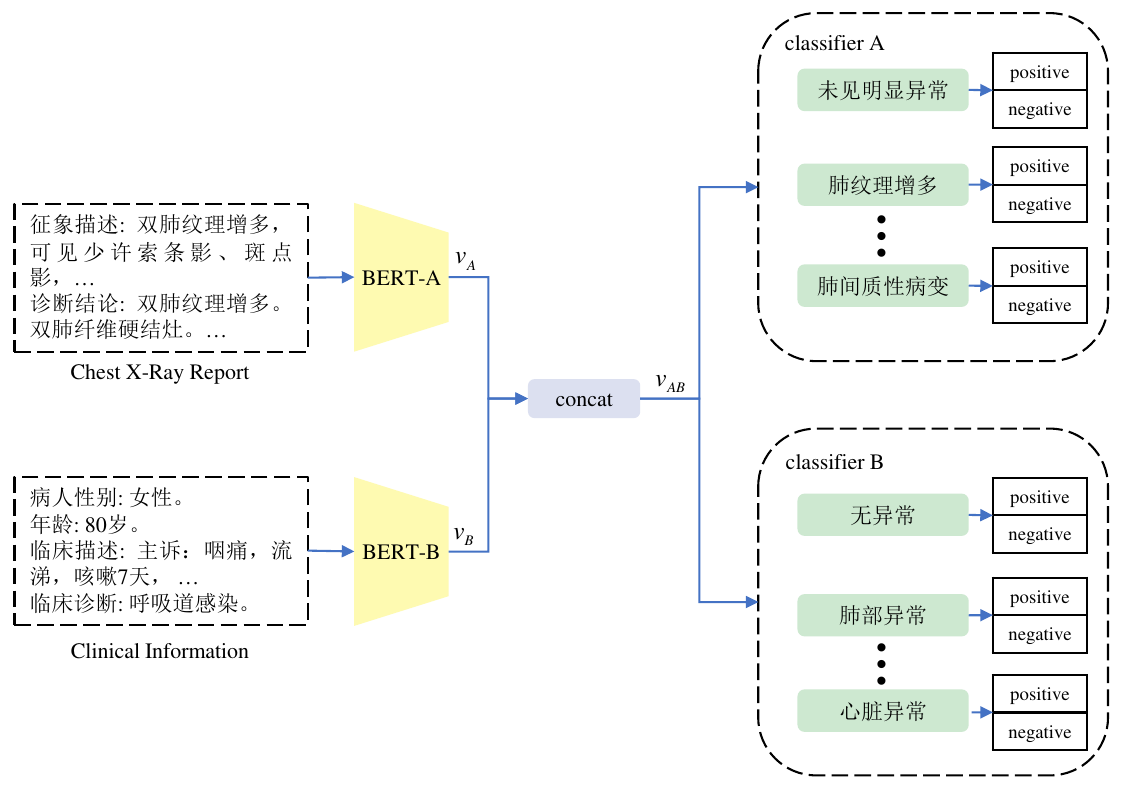} 
\caption{Overall Architecture for the Chinese Chest X-ray Report Disease Labeler} 
\label{fig:architecture} 
\end{figure}

The overall architecture of the proposed Chinese chest X-ray report disease label labeler is depicted in Figure \ref{fig:architecture}. Initially, BERT-A is tasked with encoding the medical report to obtain feature vectors $v_A$, while BERT-B encodes clinical information to acquire corresponding features $v_B$. Subsequently, the feature vectors from both are concatenated to obtain a comprehensive textual feature $v_{AB}$. Finally, these features are fed into classifier A and classifier B for supervised training or for the annotation of disease labels.  

The introduction of the dual BERT encoders stems from previous research that typically only used medical reports as model input, neglecting related clinical information. In this paper, separate dual BERT architecture are employed to encode medical reports and clinical information independently. Due to the significant content differences between the two, a weight-sharing mechanism is not adopted. Instead, each BERT is allowed to have its own weight parameters, enabling better handling of their respective textual data.   

\subsection{Hierarchical Label Learning} 

\begin{figure}[htb]
\centering 
\includegraphics[width=0.8\textwidth]{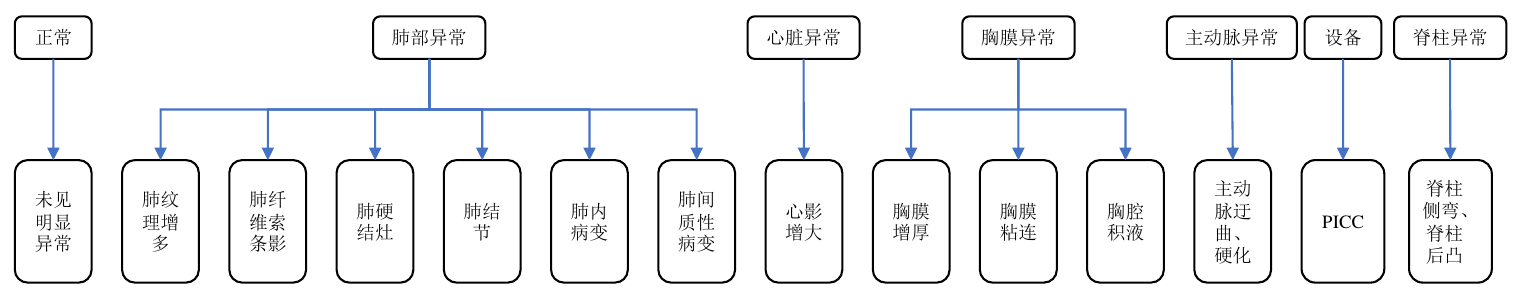} 
\caption{Hierarchical Labels Relationship Diagram} 
\label{fig:relationship} 
\end{figure}

The design of the hierarchical label learning algorithm takes into account that patients may suffer from multiple diseases simultaneously, and these diseases often exhibit correlations, with the location of the disease serving as a connection point for such correlations. Based on this, this paper has designed a hierarchical labels relationship graph to reflect the affiliation between diseases and body parts, as shown in Figure \ref{fig:relationship}. For example, if a patient is annotated with the secondary disease label "肺结节", then accordingly, the primary disease label "肺部异常" should also be marked as positive. Similarly, if a patient is labeled with the secondary disease tag "阴影增大", then the primary label "心脏异常" should also be marked as positive. If a patient is not annotated with any primary disease label, they should be classified as "正常". During the training phase of the labeler, Classifier A uses secondary labels for supervised training, while Classifier B uses primary labels. In the inference phase, only the output results of Classifier A are needed. According to the design of the hierarchical labels, Classifier A consists of 14 linear output heads, with 12 corresponding to various medical diseases, one corresponding to "PICC," and another corresponding to "未见明显异常". Classifier B consists of 7 linear output heads, with 5 corresponding to abnormal locations, one corresponding to "正常", and another corresponding to "设备".   

\section{Construction of a Chinese Chest X-ray Report Dataset} 

\subsection{Chest X-ray Image Preprocessing} 

During the preprocessing stage of constructing a Chinese chest X-ray image report dataset, the collection of chest X-ray images originated from the Image Archiving and Communication System (PACS) of the Third Hospital of Peking University. These images encompass various scenarios, including inpatient, outpatient, and health examination settings. The original images were in the Digital Imaging and Communications in Medicine (DICOM) standard format, which contained patient personal information and detailed examination-related information. In the image selection process, chest X-ray from patients under 18 years old, occupational pneumoconiosis chest X-ray, bedside chest X-ray (due to poor imaging quality), reports with irregular descriptions or overly brief content, and rib series chest X-ray were excluded. To protect patient privacy, de-identification was performed when exporting images from the system, and the hospital’s radiology information system was used to associate the images with corresponding radiological reports. This study only included chest X-ray in the posteroanterior (PA) and lateral (LA) projection directions.  

\setcounter{footnote}{0} 
After obtaining the de-identified DICOM original images, a format conversion was carried out to enable display on standard devices, with PNG format selected over JPEG because PNG can preserve more medical image details without loss. The first step in the conversion process was to read the DICOM images using the pydicom library\footnote{https://zenodo.org/records/4248192}. Subsequent to that, the Window Width and Window Center attribute values of the images were obtained. The concept of window width and window level arises because medical images typically use 16 bits to represent the information of each pixel, whereas standard monitors can only display 8-bit data, i.e., grayscale values from 0 to 255. Thus, window width and window level are used to determine the range of grayscale of interest for appropriate display on monitors. The window level specifies the grayscale center, and the window width determines the display range from darkest to brightest. A larger window width results in a smoother transition from black to white in the image but may make it more difficult to discern details; conversely, a smaller window width increases image contrast, making certain details more prominent but potentially causing the loss of other details. Therefore, when converting from DICOM to PNG format, a pixel value mapping transformation is required, mapping the values within window level ± window width/2 to the 0-255 grayscale range. The specific formula for the mapping is as follows:   

\begin{equation}
G V= 
\begin{cases}
0, & \text { if } P V<W C-\frac{W W}{2} \\ 
\frac{255 *\left(P V-\left(W C-\frac{W W}{2}\right)\right)}{W W}, & \text { if } W C-\frac{W W}{2} \leq P V \leq W C+\frac{W W}{2} \\ 
255, & \text { if } P V>W C+\frac{W W}{2}\end{cases}
\end{equation}

wherein, $GV \in [0,255]$ represents the grayscale value after mapping, $PV$ is the original pixel value, $WC$ stands for Window Center value, and $WW$ stands for Window Width value. This study directly utilized the dcmj2pnm tool from the DCMTK toolkit\footnote{https://github.com/DCMTK/dcmtk} to complete the pixel value mapping according to the formula above. The corresponding terminal command format is: "dcmj2pnm +Ww [window center value] [window width value] +on [original file path] [save file path]".    

It should be noted that for DICOM images with multiple sets of window width and window level values, this study by default chooses the first set of parameters for the image format conversion.

\subsection{Chest X-ray Report Preprocessing}
In the process of constructing Chinese chest X-ray report dataset, the preprocessing of chest X-ray reports is crucial. These reports are structured, with the radiologists' diagnoses and relevant patient information stored in specific fields. A typical radiological report, as shown in Table \ref{tab:field}, contains multiple fields: the ACC field serves as a unique identifier for the patient's current examination; the findings field records the radiologist's observational descriptions of the images; the impression field contains the doctor's diagnosis based on the radiological observations; the clinical diagnosis field records the clinical doctor's diagnosis; and the clinical description field provides relevant information about the patient.  

\begin{table}[h]
\centering
\caption{Examples of Content for Each Field in Chest X-ray Reports}
\label{tab:field}
\begin{tabular}{cccc} 
\hline
\textbf{Field Name}  & \textbf{Content}  \\
\hline
ACC  & 01220110301300 \\
\hline
征象描述 & 
\makecell{对比2021-03-23日片：双肺纹理增多、紊乱，见多发网格影，\\
左下肺新发条片样密度增高模糊影，双肺下野见点状高密度影，\\
肺门影不大，纵隔不宽，心影饱满，两膈光滑，肋膈角锐利。\\
双侧顶部胸膜增厚。余大致同前。左肾可见插管影。} \\
\hline
诊断结论 & \makecell{双肺间质性病变伴左下肺感染？较前进展，随诊 \\
双肺结节，随诊 \\
双下肺纤维硬结灶可能 \\
双侧顶部胸膜增厚}
 \\
 \hline
临床诊断& 肾造瘘术后，左 \\
\hline
病人性别 & 男 \\
\hline
年龄& 082Y00M20D \\
\hline
临床描述& \makecell{放射科号:/身高(cm):/体重(kg):/是否肝肾功能不全:\\
/是否碘剂过敏://入院检查} \\
\hline
\end{tabular}
\end{table}

In this study, the preprocessing of report data is divided into two parts: text preprocessing and annotation of medical report disease labels. Text preprocessing is mainly achieved through Python programming and regular expressions, with the relevant regular expressions shown in Table \ref{tab:re}. For example, items 1, 2, and 6 involve comparative information between the current chest X-ray and previous examinations. Since only the current chest X-ray is input during model training, and previous chest X-ray information is not available, it was decided to remove this comparative information to avoid generating errors. Items 3, 4, and 5 contain subjective opinions of radiologists, which may vary from doctor to doctor and result in different descriptions for the same chest X-ray. Therefore, to reduce data noise, these contents were removed. Item 7 aims to clear noise information from the clinical description field. Additionally, all English punctuation marks have been replaced with Chinese punctuation.  

\begin{table}[h]
\centering
\caption{Regular Expressions Used for Text Preprocessing} 
\label{tab:re}
\begin{tabular}{cccc} 
\hline
\textbf{Item No.}  & \textbf{Regular Expression} & \textbf{Matched Text}  \\
\hline
1  & \makecell{(，|。)*(余大致同前|大致同前|似大致同前| \\
余所见大致同前|所见大致同前|范围大致同前)} & ，大致同前 \\
\hline
2 & 
\makecell{\texttt{(对比|与|结合)(上片|前片)?} \\
\texttt{\textbackslash d\{3,4\}(-|.)\textbackslash d\{1,2\}(-|.)\textbackslash d\{1,2\}(日|\textbackslash s)?} \\
\texttt{(\textbackslash d\{1,2\}(：|:)\textbackslash d\{1,2\})?} \\
\texttt{(片对比|片|胸片|床旁片|床旁平片|床旁胸片|CT)?} \\
\texttt{(：|:|。|，|；|;)?}} & 对比2021-03-23日片：\\
\hline
3 & \makecell{(，|。)?(余|建议|请|清|位置)?结合.*?(。|，)} & ，请结合CT检查。 \\
 \hline
4& (，|。|、)?随诊.*?(。|，|、) &  ，随诊复查。 \\
\hline
5 & ，请?注意心功能  & ，请注意心功能 \\
\hline
6& \makecell{(，|、)?(范围|左肺|右肺|左侧|右侧|右肺野|左肺野)? \\
较前(明显|稍|略|有所)?(好转|吸收|减轻|进展|增大| \\
减少|减小|缩小|增多|改善|复张|增多|加重|增加|好转|清晰)} & ，较前稍减轻\\
\hline
7& \makecell{放射科号:/身高(cm):/体重(kg):/是否肝肾功能不全:\\
是否碘剂过敏:/*}  & \makecell{放射科号:/身高(cm):/体重(kg): \\
/是否肝肾功能不全:/是否碘剂过敏://}\\
\hline
\end{tabular}
\end{table}

Regarding the processing of specific fields, each disease diagnosis in the diagnostic conclusion field is treated as a phrase and connected with periods. The age field extracts only the first three characters and converts them to an integer type. 

Subsequently, the disease label annotation of the radiological reports is completed using the labeler proposed in Section 3 "Chinese Chest X-Ray Report Disease labeler" of this paper. The model's input includes image descriptions, diagnostic conclusions, patient gender, age, clinical descriptions, and clinical diagnoses, while the output of the model has selected 14 disease labels based on the number of disease samples and their degree of importance. These labels include "未见明显异常", "肺纹理增多", "肺纤维索条影", "心影增大", "肺硬结灶", "胸膜增厚", "主动脉迂曲、硬化", "PICC," "肺结节", "肺内病变", "胸膜粘连", "脊柱侧弯、脊柱后凸", "胸腔积液", and "肺间质性病变". Through the steps mentioned above, the preprocessing of the radiological reports and the generation of disease labels were completed. 

\subsection{Overview of the Constructed CCXRD Dataset}

The Chinese chest X-ray report dataset constructed in this paper is named CCXRD (Chinese Chest X-ray Report Dataset). The dataset is randomly divided according to an 8:1:1 ratio of data to form training, validation, and test sets. As shown in Table \ref{tab:distribution}, it includes detailed statistical information on the medical imaging of the three datasets. PA images refer to chest X-ray taken in the anteroposterior projection, while LA images are taken in the lateral projection. The data division is based on the number of PA images, as each sample has only one PA image and one medical text report, while LA images are not mandatory. Accordingly, the Chinese chest X-ray image report dataset constructed in this study contains a total of 47,886 samples, with a total of 51,262 images. Among these 51,262 images, 33,172 images show at least one disease, and 18,060 images are normal chest X-ray.  

\begin{table}[h]
\centering
\caption{Overview of Image Distribution Across Training, Validation, and Test Sets} 
\label{tab:distribution}
\begin{tabular}{ccccc} 
\hline
\textbf{Dataset}  & \textbf{Training set} & \textbf{Validation set} & \textbf{Test set} & \textbf{total} \\
\hline
Number of images& 40,968 & 5,156 & 5,138 & 51,262 \\
Number of PA images& 38,308 & 4,789& 4,789 & 47,886 \\
Number of LA images & 2,660 & 367	 & 349	 & 3,376 \\
\hline
\end{tabular}
\end{table}

A statistical analysis of the medical report disease labels in the dataset has been conducted, with the results presented in Table \ref{tab:statistics}. The positive ratio is calculated by dividing the number of positive samples by the total number of samples in the dataset; therefore, the sum of positive ratios may exceed 100\%, which is due to the presence of samples annotated with multiple disease labels. In the dataset, "未见明显异常" and "肺纹理增多" are the most numerous, accounting for 34.9\% and 48.93\% of the samples, respectively, a figure that results from a reduced proportion of normal samples. In the actual distribution of medical data, the proportion of normal chest X-ray might be higher. Moreover, common clinical diagnoses such as "肺纤维索条影", "心影增大", "肺硬结灶", and "胸膜增厚", while lower in proportion compared to normal samples, are still relatively common in clinical practice. The proportions of other diseases are smaller, which poses a significant challenge in the training of related models and also reflects a distinctive feature of this dataset.  

\begin{table}[h]
\centering
\caption{Statistics of Disease Label Counts and Their Frequencies in the Dataset} 
\label{tab:statistics}
\begin{tabular}{ccccc} 
\hline
\textbf{Disease Name}  & \textbf{Positive Quantity} & \textbf{Positive Ratio}  \\
\hline
未见明显异常 & 16,714 & 34.90\% \\
肺纹理增多 & 23,429 & 48.93\% \\
肺纤维索条影 & 7,792 & 16.27\% \\
心影增大 & 5,722 & 11.95\% \\
肺硬结灶 & 6,181 & 12.91\% \\
胸膜增厚 & 4,164 & 8.70\% \\
主动脉迂曲、硬化 & 1,502 & 3.14\% \\
PICC & 1,077 & 2.25\% \\
肺结节 & 1,634 & 3.41\% \\
肺内病变 & 1,208 & 2.52\% \\
胸膜黏连 & 909 & 1.90\% \\
脊柱侧弯、脊柱后凸 & 688 & 1.44\% \\
胸腔积液 & 1,373 & 2.87\% \\
肺间质性病变 & 1,384 & 2.89\% \\
\hline
\end{tabular}
\end{table}

In summary, the CCXRD dataset proposed in this paper encompasses a total of 51,262 chest X-ray, including both frontal and lateral views, and is accompanied by 47,886 radiological reports. Each report contains fields such as "ACC," "征象描述", "诊断结论", "临床诊断", "病人性别", "年龄", "临床描述", and "疾病标签", and has been divided into training, validation, and test sets in an 8:1:1 ratio.  

\section{Experimental Results and Analysis}
        
\subsection{Validation dataset} 

To validate the effectiveness of the Chinese chest X-ray report disease labeler proposed in this paper and to construct a large-scale Chinese chest X-ray report dataset based on this labeler, this study collaborated with the Radiology Department of Peking University Third Hospital to manually annotate a data subset, which includes a total of 24,035 samples divided into a training set (19,228 samples), a validation set (2,403 samples), and a test set (2,404 samples) in an 8:1:1 ratio. The medical text data items included in the dataset cover "征象描述", "诊断结论", "患者性别", "患者年龄", "临床描述", and "临床诊断". All disease labels were personally annotated by professional diagnostic physicians, involving a total of 53 disease categories. Based on the number of samples and the clinical significance of the diseases, this study selected 14 primary disease labels for in-depth analysis.  

\subsection{Evaluation Metrics}

To comprehensively evaluate the performance of the Chinese disease labeler, this study employed multiple evaluation metrics, including F1 score, Kappa statistic\cite{cohen1960a}, weighted F1 score, and weighted Kappa statistic. Both the F1 score and Kappa statistic are key performance indicators for classification tasks, but they emphasize different aspects. 

The F1 score is defined for binary classification problems and is the harmonic mean of precision and recall. Precision represents the proportion of actual positive cases among the samples predicted as positive by the classifier, while recall indicates the proportion of actual positive cases that the classifier correctly predicted as positive. The F1 score can be calculated using the following formula: F1 = 2 × (precision × recall) / (precision + recall). The closer the F1 score is to 1, the better the model's performance. The F1 score is an important indicator for evaluating model performance on imbalanced datasets because it considers both precision and recall.

The Kappa statistic, also known as Cohen's Kappa, is another metric for assessing classifier performance, particularly suitable for multi-class problems. Unlike the F1 score, which focuses on the accuracy of classifier predictions, the Kappa statistic measures the difference between the accuracy of classifier predictions and the accuracy of random predictions, providing a corrected assessment of classifier performance. A Kappa value of 1 indicates perfect agreement between predictions and reality, a value of 0 indicates that the consistency of predictions is the same as random predictions, and a negative Kappa value suggests that the consistency of predictions is below the level of random chance. 

The weighted F1 score and weighted Kappa statistic introduce the frequency of occurrence of each disease label as a weight in the calculation process, resulting in a weighted evaluation. This weighting mechanism takes into account the distribution of different disease labels in the dataset, thus providing a more precise assessment of model performance. 

\subsection{Experimental Setup}

The dual BERT encoder used in this study is based on the BERT-Base architecture, with initial weights inherited from MedBERT-kd\cite{yang2021exploration}, and all layers were fine-tuned. In this architecture, the feature dimension of the [CLS] token output by the BERT model is 768. For the classification task, two linear classification heads were designed; classifier A has 14 output nodes, while classifier B has 7 output nodes. To prevent overfitting, the Dropout rate was set to 0.1. Regarding the loss function, Focal Loss was adopted to address the issue of severe imbalance between positive and negative samples during training. The model's optimizer was Adam, with a learning rate set to 2e-5, and the rest of the parameters remained at their default values.  

\begin{figure}[htb]
\centering 
\includegraphics[width=0.8\textwidth]{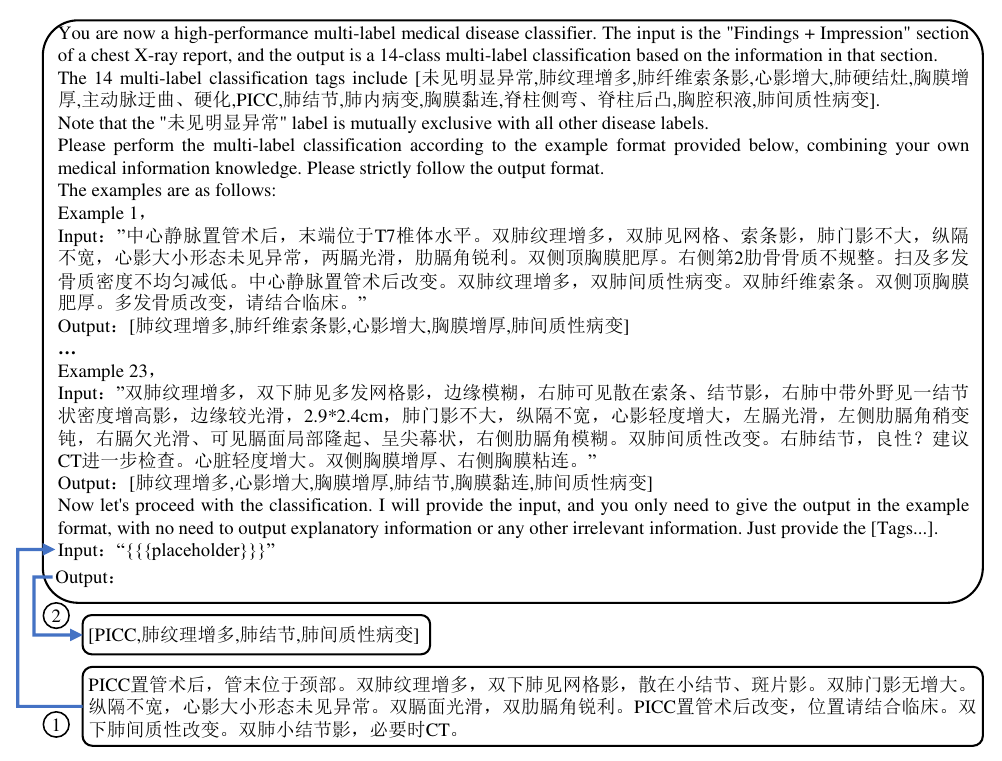} 
\caption{ChatGPT Prompts and the Corresponding Application Workflow} 
\label{fig:chatgpt} 
\end{figure}

To thoroughly assess the performance of the Chinese disease labeler, comparative experiments were designed, including the widely used CheXbert model\cite{smit2020combining} in the English medical domain, as well as the recently popular ChatGPT 3.5\footnote{https://chat.openai.com}  and ChatGPT 4\footnote{https://openai.com/gpt-4}  models. Specifically, the version for GPT-3.5 was gpt-3.5-turbo-1106, and for GPT-4 it was gpt-4-1106-preview. In the experimental setup, both GPT-3.5 and GPT-4 used the same prompts, the content of which is shown in Figure \ref{fig:chatgpt}. The example part of the prompts covered samples from all disease categories, with at least two samples included for each disease. The experimental process began by replacing the "{{{placeholder}}}" with the diagnostic report to be classified, and then inputting it into the ChatGPT model to obtain the model's classification results for the diseases.  

\subsection{Experimental Results}

The results of the comparative experiments are shown in Table \ref{tab:comparison}, where the method proposed in this study achieved the best performance across all evaluation metrics. Although the model in this paper is similar to CheXbert in terms of network architecture and training methodology, the proposed dual BERT architecture and hierarchical labels algorithm have led to superior performance. The experimental results indicate that, while the general-purpose GPT-3.5 and GPT-4 models demonstrated unexpectedly good performance, with GPT-4 showing a significant improvement in inference capability over GPT-3.5, there is still a gap compared to the method of this study. This is mainly because only one-shot learning was performed in the prompts of GPT-4. If further fine-tuning of the model were possible, a significant performance improvement could be expected, potentially even surpassing the current specialized models. However, GPT-4's fine-tuning API is not yet available.  

\begin{table}[h]
\centering
\caption{Performance Comparison of Disease Labelers}
\label{tab:comparison}
\begin{tabular}{ccccc} 
\hline
\textbf{Models}  & \textbf{F1 Score} & \textbf{weighted F1 score} & \textbf{Kappa} & \textbf{weighted Kappa}  \\
\hline
CheXbert & 0.9182 & 0.9767 & 0.9157 & 0.9730 \\
ChatGPT 3.5 & 0.8278 & 0.9047 & 0.7897 & 0.8724 \\
ChatGPT 4 & 0.9008 & 0.9260 & 0.8775 & 0.9059 \\
Our Method & \textbf{0.9430} & \textbf{0.9809} & \textbf{0.9408} & \textbf{0.9774} \\
\hline
\end{tabular}
\end{table}

In addition, to quantitatively evaluate the contribution of each proposed algorithm to model performance, this study also conducted a series of ablation experiments, the results of which are summarized in Table \ref{tab:ablation}. Here, w/o indicates the removal of the corresponding component. The results show that the removal of either the hierarchical labels algorithm or the dual BERT encoder led to a significant decrease in F1 score and Kappa statistic. Moreover, compared to the hierarchical labels algorithm, the architecture of the dual BERT encoder had a slightly weaker impact on performance, which may be due to the scarcity of effective information and the abundance of interfering information in clinical texts. 

\begin{table}[h]
\centering
\caption{Ablation Study Results for the Chinese Disease Labeler}
\label{tab:ablation}
\begin{tabular}{ccccc} 
\hline
\textbf{Models}  & \textbf{F1 Score} & \textbf{weighted F1 score} & \textbf{Kappa} & \textbf{weighted Kappa}  \\
\hline
Our method & \textbf{0.9430} & \textbf{0.9809} & \textbf{0.9408} & \textbf{0.9774} \\
w/o hierarchical labels & 0.9254 & 0.9767 & 0.9228 & 0.9737 \\
w/o dual BERT & 0.9312 & 0.9774 & 0.9288 & 0.9735 \\
\hline
\end{tabular}
\end{table}

Finally, to find the best Chinese pre-trained BERT model for initialization weights, this paper also conducted an in-depth exploratory experiment on various Chinese BERT models released over the past three years, as shown in Table \ref{tab:pretrain}. The experiment was divided into two groups: the first group of models obtained initialization weights trained on Chinese general text, while the second group was trained on Chinese medical text. The results show that for medical domain tasks, weights trained directly on medical texts are significantly superior to those trained on general texts. Although MedBERT-kd was released in 2021, and the general text-trained m3e-base and LEALLA-base were released in 2023, the performance of the latter two still did not match the former, emphasizing that sometimes the suitability of the training data can be more crucial than the training method and model architecture.  

\begin{table}[h]
\centering
\caption{Comparative Performance of Chinese Pre-trained BERT Models in Disease Classification of Medical Reports}
\label{tab:pretrain}
\begin{tabular}{ccccc} 
\hline
\textbf{Models}  & \textbf{F1 Score} & \textbf{weighted F1 score} & \textbf{Kappa} & \textbf{weighted Kappa}  \\
\hline
chinese-macbert-base\cite{cui2020revisiting} &  0.9337 & 0.9808 & 0.9318 & \textbf{0.9793} \\
RoBERTa-wwm-ext\cite{cui2021pre} &  0.9378 & 0.9804 & 0.9357 & 0.9774 \\
chinese-lert-base\cite{cui2022lert} &  0.9381 & \textbf{0.9812} & 0.9362 & 0.9778 \\
m3e-base\cite{wang2023m3e}  & 0.9387 & \textbf{0.9812} & 0.9367 & 0.9782 \\
LEALLA-base\cite{mao2023lealla} &  0.9169 & 0.9782 & 0.9146 & 0.9745 \\
\hline
FT-BERT\cite{li2020chinese} &  0.9274 & 0.9795 & 0.9253 & 0.9762 \\
MedBERT-kd\cite{yang2021exploration} &  \textbf{0.9430} & 0.9809 & \textbf{0.9408} & 0.9774 \\
\hline
\end{tabular}
\end{table}

\section{Conclusion}

This study addresses the lack of Chinese chest X-ray report disease labelers by constructing a Chinese chest X-ray report disease labeler based on a dual BERT architecture and hierarchical label learning algorithm. This labeler effectively encodes diagnostic reports and clinical information independently and leverages the hierarchical relationship between diseases and body parts to build a hierarchical label learning algorithm, significantly enhancing the accuracy of disease annotation. Subsequently, a Chinese chest X-ray report dataset (CCXRD) containing 51,262 chest X-ray samples was constructed based on this labeler. Experimental analysis conducted on a Chinese data subset built by experts verified the effectiveness of the proposed disease labeler.    

Future work will concentrate on improving the performance of the disease labeler and building a larger-scale Chinese dataset using this labeler. Amidst the growing popularity of large models, a more extensive dataset is anticipated to advance the research on automatic chest X-ray report generation, allowing it to play a vital role in clinical practice.

\bibliographystyle{unsrt}  


\end{CJK}
\end{document}